\relax
\documentclass[letterpaper]{article} 
\usepackage{aaai21}  
\usepackage{times}  
\usepackage{helvet} 
\usepackage{courier}  
\usepackage[hyphens]{url}  
\usepackage{graphicx} 
\usepackage{algorithm}
\usepackage{algorithmic}
\usepackage{amssymb}
\usepackage{amsmath}
\usepackage[backref]{hyperref} 

\usepackage{epsfig}
\usepackage{graphicx}
\usepackage{multirow}
\usepackage{tabularx}
\usepackage{booktabs}
\usepackage{url}
\usepackage{makecell}

\usepackage{float}
\usepackage{stfloats}
\usepackage{subfigure}
\usepackage{graphicx}

\usepackage{natbib}  
\usepackage{caption} 
\title{PP-OCRv3: More Attempts for the Improvement of \\ Ultra Lightweight OCR System}
\author {
    Chenxia Li, Weiwei Liu, Ruoyu Guo, Xiaoting Yin, Kaitao Jiang,  Yongkun Du, \\
   Yuning Du, Lingfeng Zhu, Baohua Lai, Xiaoguang Hu, Dianhai Yu, Yanjun Ma \\
}
\affiliations {
    Baidu Inc. \\
    \{lichenxia, zhulingfeng\}@baidu.com
}

\begin{document}

\maketitle

\begin{abstract}
Optical character recognition (OCR) technology has been widely used in various scenarios, as shown in Figure \ref{vis_sys}. Designing a practical OCR system is still a meaningful but challenging task. In previous work, considering the efficiency and accuracy, we proposed a practical ultra lightweight OCR system (PP-OCR), and an optimized version PP-OCRv2. In order to further improve the performance of PP-OCRv2, a more robust OCR system PP-OCRv3 is proposed in this paper. PP-OCRv3 upgrades the text detection model and text recognition model in 9 aspects based on PP-OCRv2. For text detector, we introduce a PAN module with large receptive field named LK-PAN, a FPN module with residual attention mechanism named RSE-FPN, and DML distillation strategy. For text recognizer, we introduce lightweight text recognition network SVTR-LCNet, guided training of CTC by attention, data augmentation strategy TextConAug, better pre-trained model by self-supervised TextRotNet, U-DML, and UIM to accelerate the model and improve the effectiveness. Experiments show that Hmean of PP-OCRv3 outperforms PP-OCRv2 by 5\% with comparable inference speed. All the above mentioned models are open-sourced and the code is available in the GitHub repository PaddleOCR \footnote{https://github.com/PaddlePaddle/PaddleOCR} which is powered by PaddlePaddle \footnote{https://github.com/PaddlePaddle}.
\end{abstract}

\section{Introduction}
\begin{figure}[t]
\centering
\subfigure{
\begin{minipage}[t]{\linewidth}
\centering
\includegraphics[width=\columnwidth]{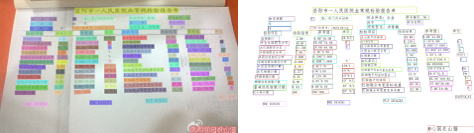}
\end{minipage}
}

\subfigure{
\begin{minipage}[t]{\linewidth}
\centering
\includegraphics[width=\columnwidth]{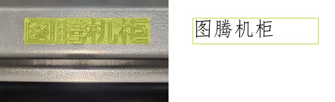}
\end{minipage}
}

\caption{Some image results of the proposed PP-OCRv3 system.}
\label{vis_sys}
\end{figure}

OCR (Optical Character Recognition) in the wild, as shown in Figure \ref{vis_sys}, has various applications scenarios, such as document electronization, identity authentication, digital financial system, and vehicle license plate recognition. In recent years, researchers have conducted in-depth research on the sub problems of text detection and text recognition in OCR. Many effective algorithms have been proposed, such as DB \cite{liao2020real} for text detection and CRNN \cite{shi2016end} for text recognition. By connecting a detection model and a recognition model, a common two-stage OCR system can be obtained. In practical industrial applications, OCR systems often need to be deployed in various software and hardware environments, where the storage space or computing resources are often limited, such as a mobile phone. Therefore, it is necessary to consider both the accuracy and the computational efficiency when we build an OCR system in practical.

Previously, we proposed a practical ultra lightweight OCR system (PP-OCR) \cite{du2020pp} to balance the accuracy against the efficiency and a bag of tricks in PP-OCRv2 \cite{du2021pp} to further improve the accuracy without increasing prediction cost. The Hmean of PP-OCRv2 is 7\% higher than that of PP-OCR mobile models with the same prediction cost and is comparable to the server models which use ResNet series as backbones. However, there are still some badcases such as missed detection of single word, misrecognition, as shown in Figure \ref{vis_det} and Figure \ref{rec_examples}. In this paper, we propose PP-OCRv3, which is a more robust OCR system, and can better figure the mentioned problems. 

\begin{figure*}[t]
\centering
\includegraphics[width=15cm]{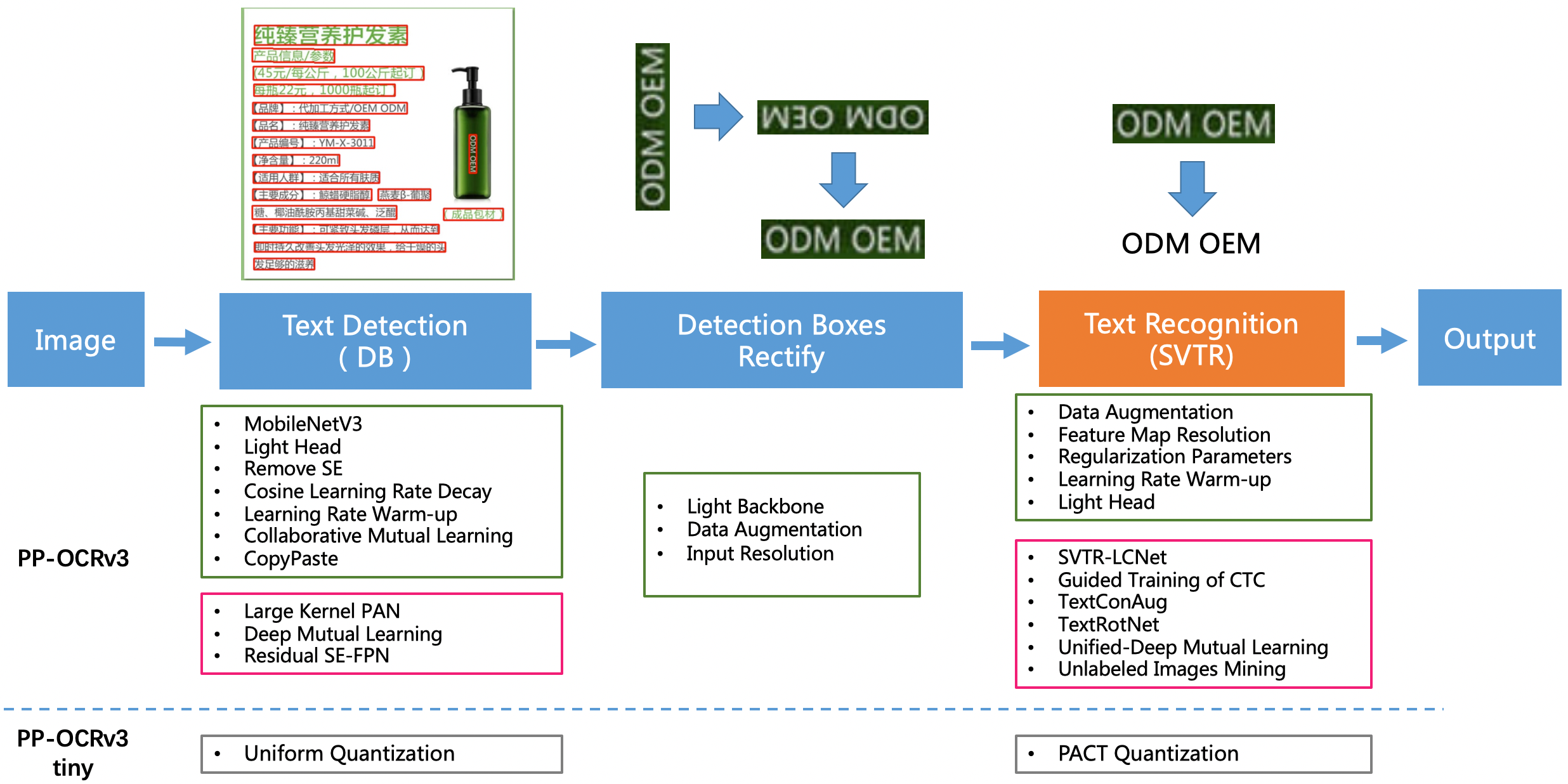}
\caption{Framework of the proposed PP-OCRv3. Strategies in the green boxes are the same as PP-OCRv2. Strategies in the pink boxes are the newly added ones in the PP-OCRv3. Strategies in the gray boxes are adopted by PP-OCRv3 tiny models.}
\label{framework}
\end{figure*}

Figure \ref{framework} illustrates the framework of PP-OCRv3. The overall framework of PP-OCRv3 is the same as that of PP-OCRv2, which consists of three parts: text detection, detected boxes rectification and text recognition. In PP-OCRv3, the text detection model and text recognition model are further optimized, respectively. Specifically, the detection network is still optimized based on DB \cite{liao2020real} , while the base model of recognition network is replaced by SVTR \cite{Du2022SVTRST} instead of CRNN \cite{shi2016end}.

Most strategies follow PP-OCR and PP-OCRv2 as shown in the green boxes. The strategies in the pink boxes are newly proposed in PP-OCRv3. 

For text detector, we introduce a PAN module with large receptive field named LK-PAN, a FPN module with residual attention mechanism named RSE-FPN, and DML \cite{dml2017} distillation strategy. LK-PAN and DML are used to improve the performance of the teacher model, while RSE-FPN is integrated into the student network. With a better performed teacher model and an optimised student network, a better detection model can be trained with Collaborative Mutual Learning (CML) \cite{du2021pp} . 

For text recognizer, we introduce lightweight text recognition network SVTR-LCNet, guided training of CTC by attention, data augmentation strategy TextConAug, better pre-trained model by self-supervised TextRotNet, U-DML, and UIM to achieve the balance of accuracy and efficiency. SVTR-LCNet is a novel lightweight text recognition network which combines the Transformer-based algorithm SVTR \cite{Du2022SVTRST} and the CNN-based algorithm PP-LCNet \cite{cui2021pplcnet} , which was utilized as the backbone of the recognizer of PP-OCRv2, so as to combine their advantages in accuracy and efficiency. The other six strategies, including guided training of CTC by attention, TextConAug, TextRotNet, U-DML, and UIM are introduced to improve the accuracy without increasing any prediction cost.

Besides, the strategies in the gray boxes of Figure \ref{framework} are adopted to further compress and speed up PP-OCRv3. The compressed model is named PP-OCRv3 tiny. 

We conduct a series of ablation experiments to verify the effectiveness of the above strategies. Experiments show that Hmean of PP-OCRv3 outperforms PP-OCRv2 by 5\% with comparable inference speed.

The rest of the paper is organized as follows. In section 2, we present the details of the newly proposed improvement strategies. Experimental results are discussed in section 3 and conclusions are conducted in section 4.

\section{Improvement Strategies}
 
\subsection{Text Detection}

\begin{figure*}[tb]
\centering
\subfigure{
\centering
\includegraphics[width=0.8\textwidth]{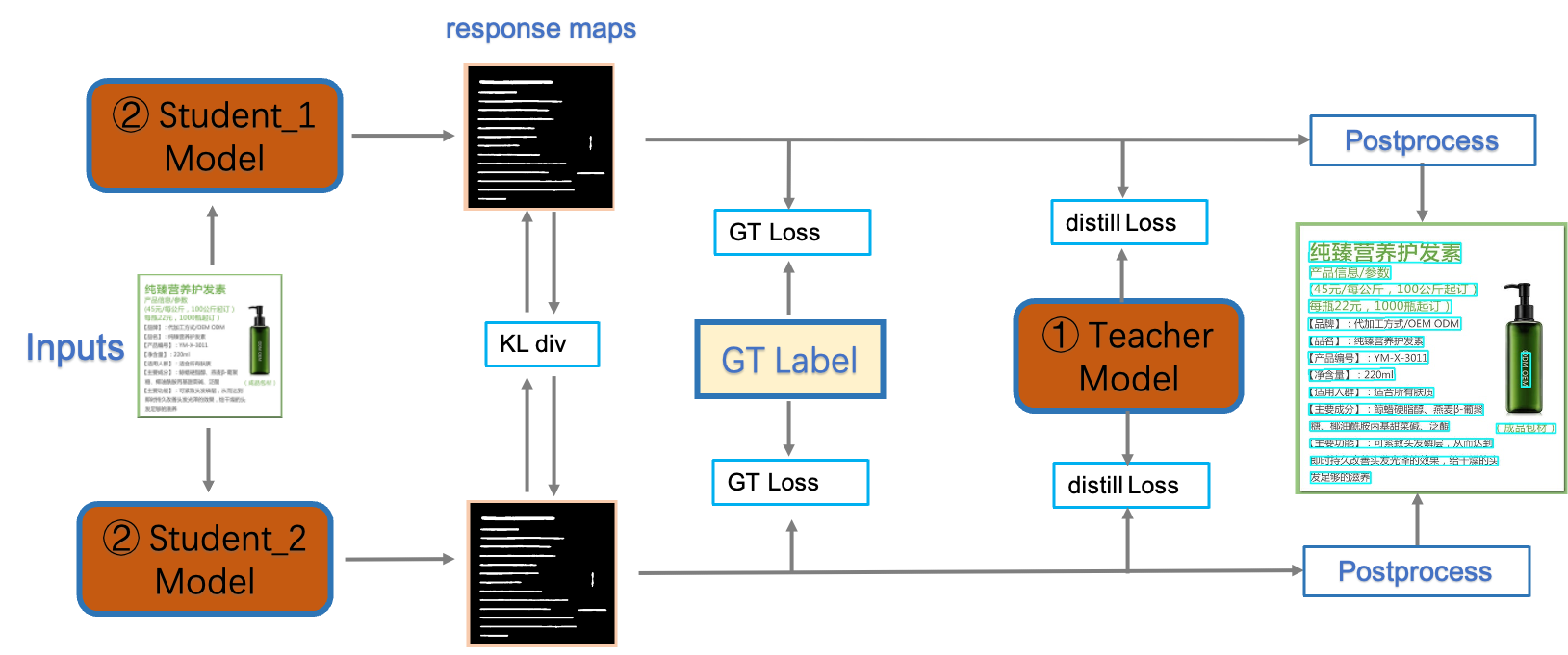}
}
\caption{CML distillation framework of PP-OCRv3 detection model.}
\label{CML}
\end{figure*}

The training framework of PP-OCRv3 detection model is still CML (Collaborative Mutual Learning) distillation, which was proposed in PP-OCRv2, as shown in Figure \ref{CML}. The main idea of CML is to combine the traditional distillation strategy of Teacher guiding Student and DML(Deep Mutual Learning) \cite{zhang2018deep}, which allows the Student networks to learn from each other. In PP-OCRv3, we optimize the teacher model and the student model respectively. For the teacher model, a PAN module with large receptive field named LK-PAN is proposed and the DML distillation strategy is adopted; for the student model, a FPN module with residual attention mechanism named RSE-FPN is proposed.

\subsubsection{LK-PAN: A PAN module with large receptive field}
\begin{figure}[tb]
\centering
\subfigure{
\centering
\includegraphics[width=\columnwidth]{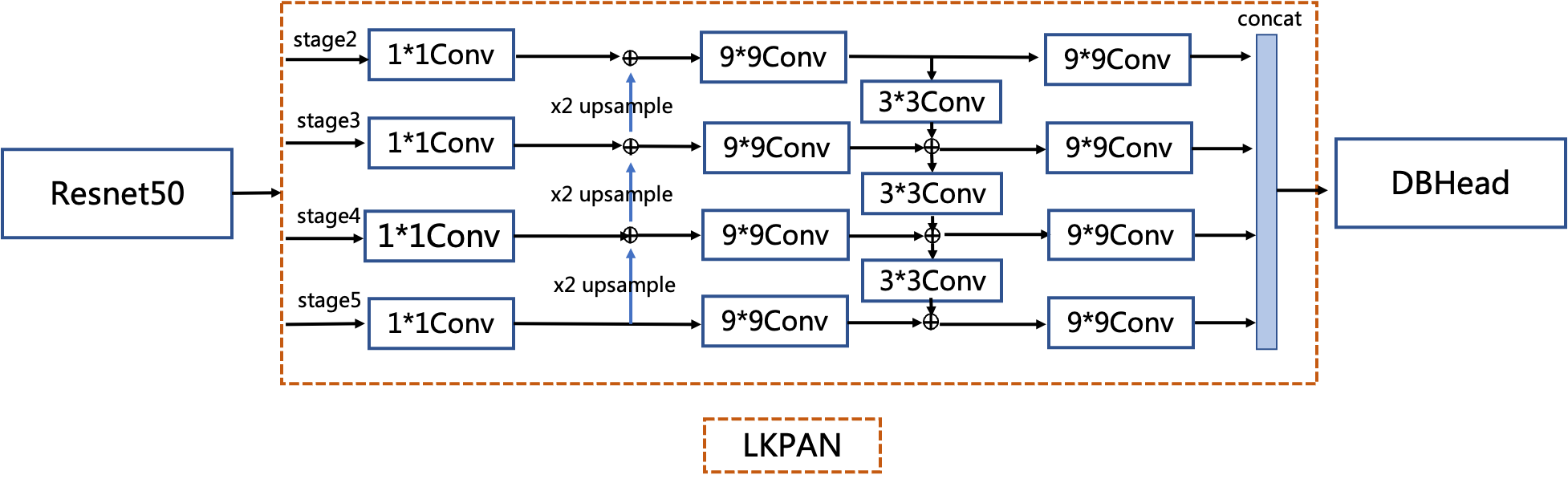}
}
\caption{Schematic diagram of LK-PAN.}
\label{LKPAN}
\end{figure}

LK-PAN (Large Kernel PAN) is a lightweight PAN \cite{liu2018path} module with larger receptive field as shown in Figure \ref{LKPAN}. The main idea is to increase the convolution kernel size in the path augmentation of the PAN module from $3 \times 3$ to $9 \times 9$, which can improve the receptive field of each pixel of the feature map, making it easier to detect text in large fonts and text with extreme aspect ratios.

\subsubsection{DML: Deep Mutual Learning for Teacher Model} 
\begin{figure}[tb]
\centering
\subfigure{
\centering
\includegraphics[width=\columnwidth]{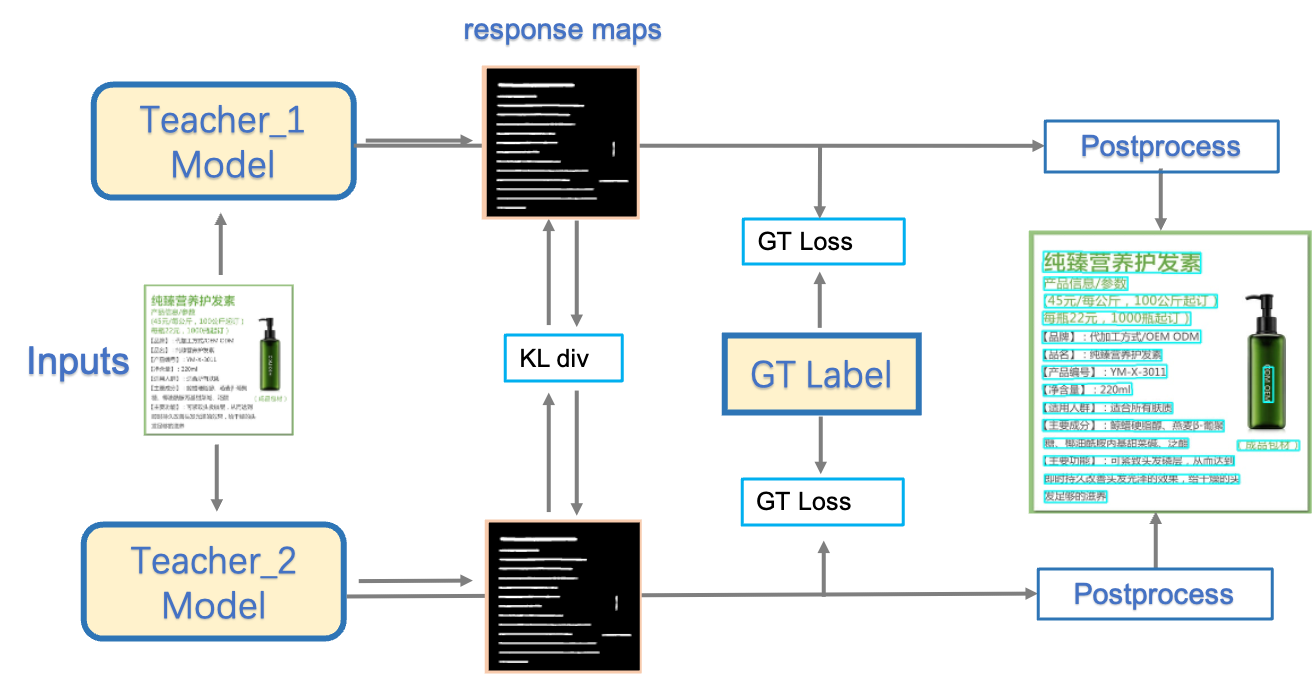}
}
\caption{Schematic diagram of DML.}
\label{DML}
\end{figure}

DML(Deep Mutual Learning) \cite{zhang2018deep} can effectively improve the accuracy of the text detection model by learning from each other with two models with the same structure. The DML strategy is adopted in the teacher model training to improve the Hmean of the teacher model as much as possible. The schematic diagram of DML in PP-OCRv3 is shown in Figure \ref{DML}.

\subsubsection{RSE-FPN: A FPN module with residual attention mechanism} 
\begin{figure}[tb]
\centering
\subfigure{
\centering
\includegraphics[width=\columnwidth]{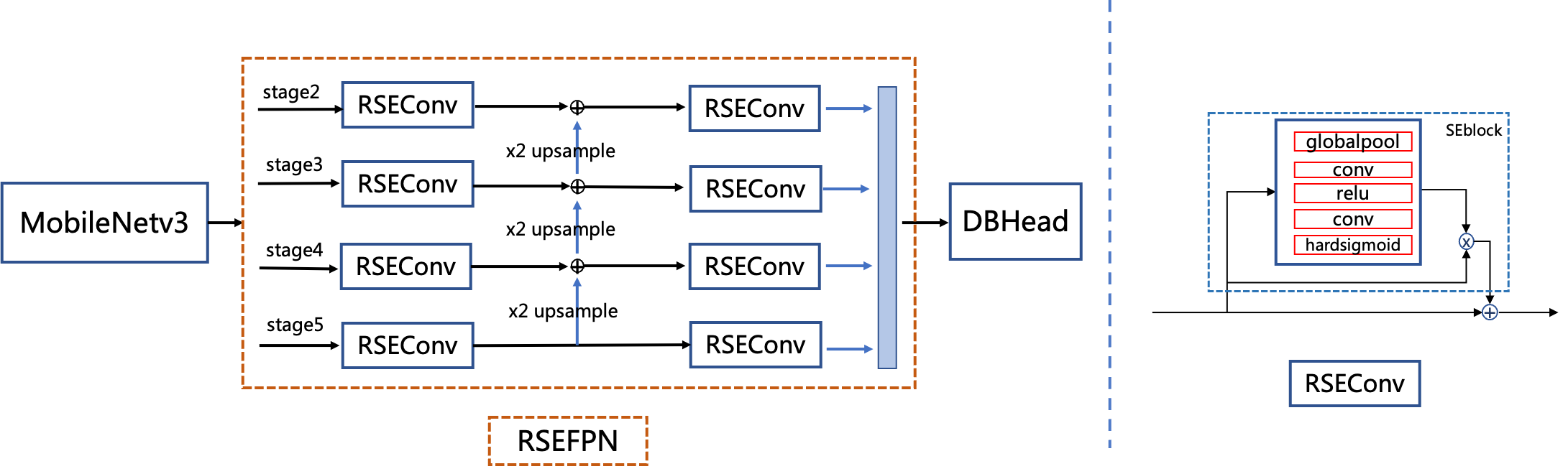}
}
\caption{Schematic diagram of RSE-FPN.}
\label{RSEFPN}
\end{figure}

RSE-FPN (Residual Squeeze-and-Excitation FPN) introduces residual attention mechanism by replacing the convolution layers in FPN with RSEConv, to improve the representation ability of the feature map. RSEConv consists of two parts: Squeeze-and-Excitation(SE) block \cite{hu2018squeeze} and the residual structure, as shown in Figure \ref{RSEFPN}. At first, we tried to add only SE blocks, which turned out not as effective as expected. Considering the number of channels of the lightweight FPN of PP-OCRv2 is relatively small, the SE module may suppress some channels containing important features. The introduction of residual structure in RSEConv can alleviate the above problems and improve the text detection performance. 

\subsection{Text Recognition}
\begin{figure*}[ht]
\centering
\subfigure{
\centering
\includegraphics[width=0.8\textwidth]{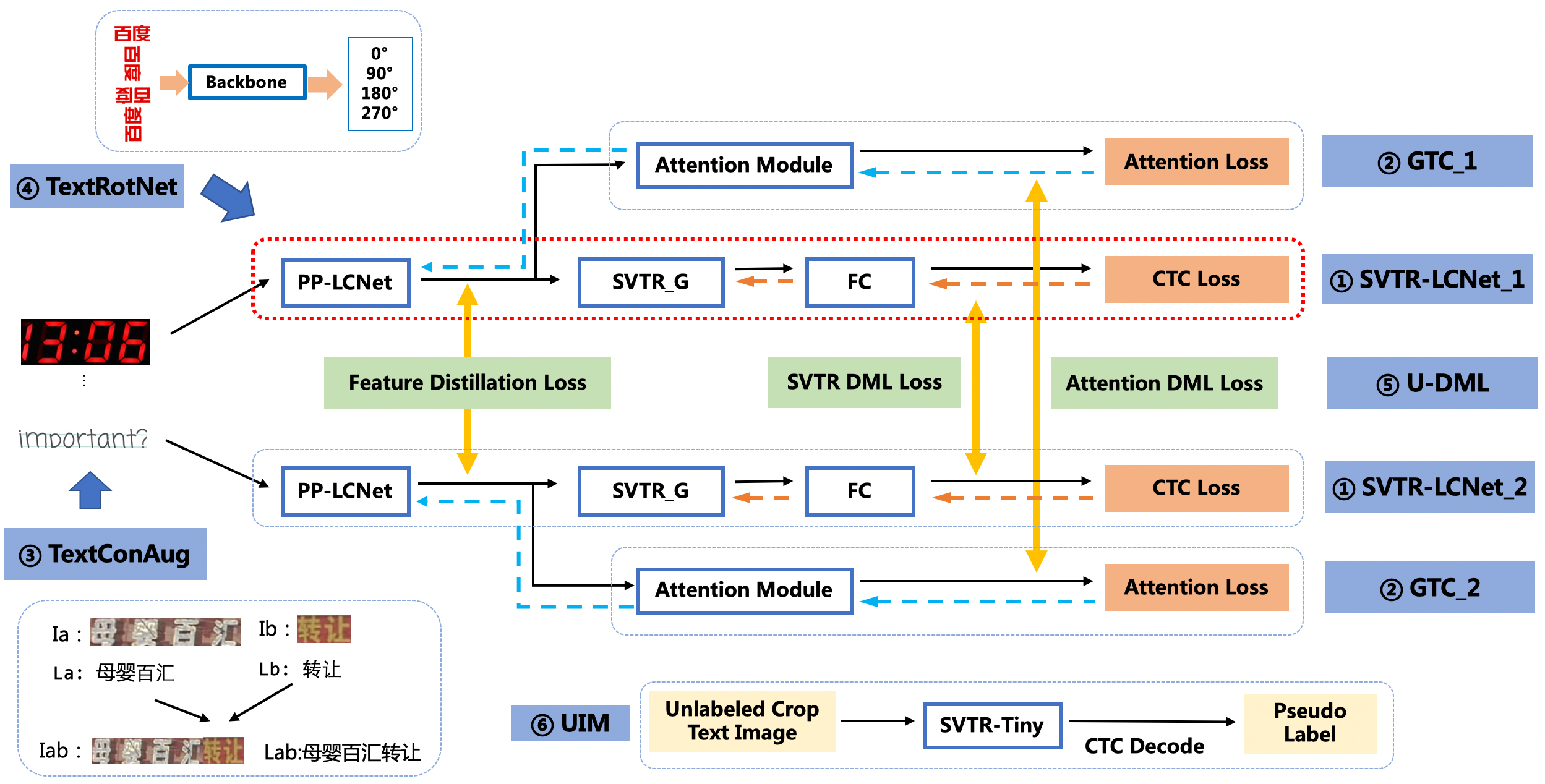}
}
\caption{Framework and training process of PP-OCRv3 recognition model.}
\label{ppocrv3_rec}
\end{figure*}

The recognition model of PP-OCRv3 is optimized based on the text recognition algorithm SVTR \cite{Du2022SVTRST}. SVTR no longer involves RNN(Recurrent Neural Network) by introducing transformers structure, which can mine the context information of text line image more effectively. To make SVTR more practical, we adopt six strategies to optimize and accelerate the model, as shown in Figure \ref{ppocrv3_rec}.

\subsubsection{SVTR-LCNet: Lightweight Text Recognition Network}
\begin{figure}[tb]
\centering
\subfigure{
\centering
\includegraphics[width=\columnwidth]{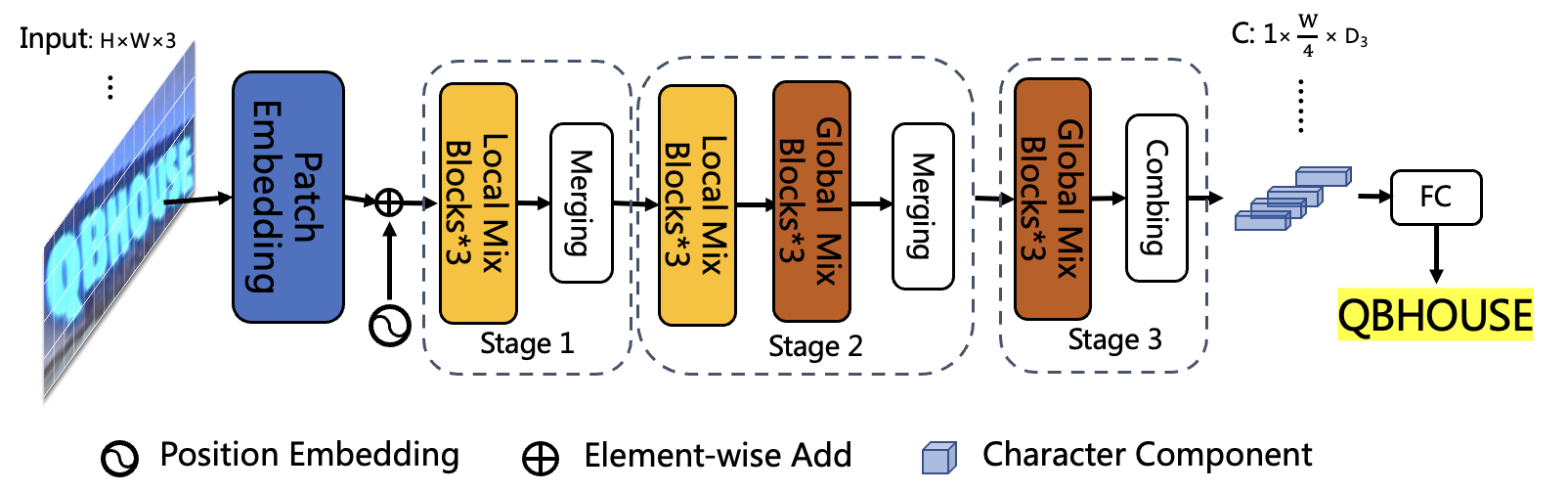}
}
\caption{Framework of the SVTR-Tiny.}
\label{SVTR_Tiny}
\end{figure}

\begin{figure}[h!]
\centering
\subfigure[Replace the first half network of SVTR  with the first three stages of PP-LCNet, and retain only 4 Global Mix Blocks.]{
    \begin{minipage}{8cm}
        \centering
        \includegraphics[width=\columnwidth]{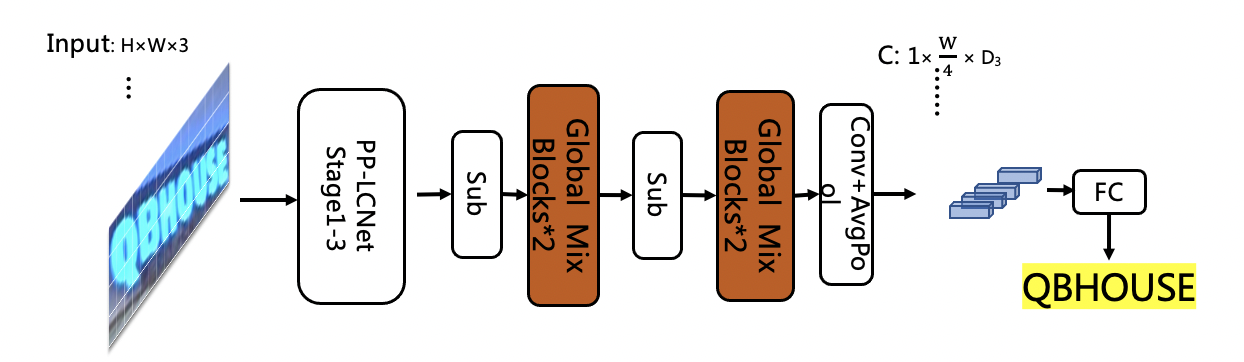}
    \end{minipage}
}

\subfigure[Reduce the number of Global Mix Blocks from 4 to 2.]{
    \begin{minipage}{8cm}
        \centering
        \includegraphics[width=\columnwidth]{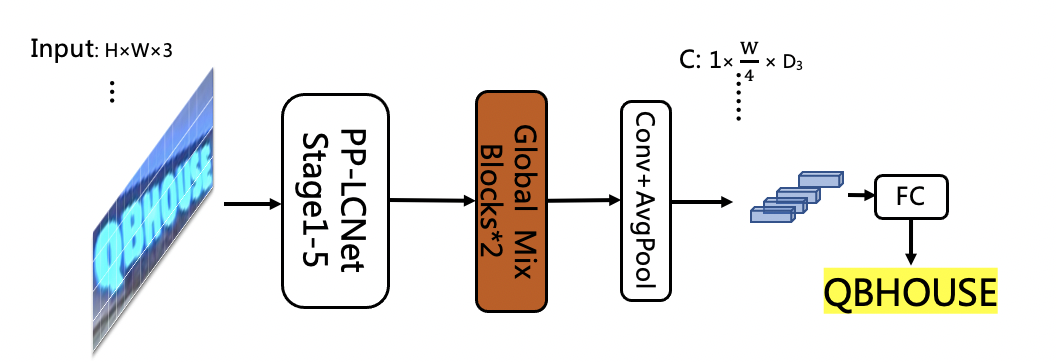}
    \end{minipage}
}

\subfigure[Move the Global Mix Block behind pooling layer. The dotted line represents the position movement process of the Global Mix Blocks which is denoted as SVTR\_G.]{
    \begin{minipage}{8cm}
        \centering
        \includegraphics[width=\columnwidth]{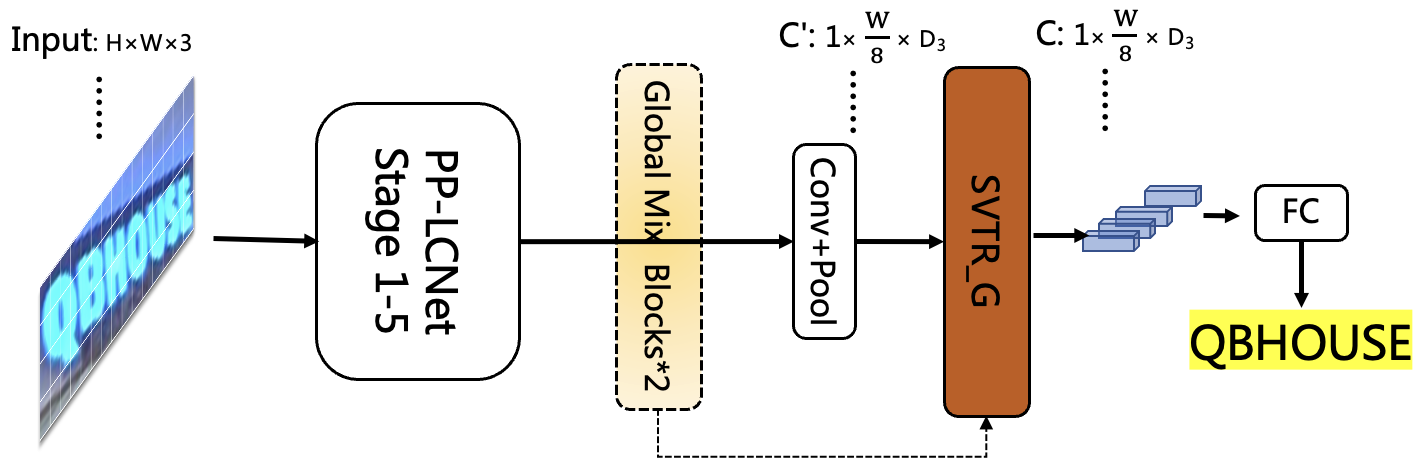}
    \end{minipage}
}
\caption{Optimization process of SVTR-Tiny. SVTR-LCNet is shown as (c).}
\label{svtr_modify_process}
\end{figure}

SVTR-LCNet is a lightweight text recognition network fusing Transformer-based network SVTR \cite{Du2022SVTRST} and lightweight CNN-based network PP-LCNet \cite{cui2021pplcnet}. Specifically, we adopt a tiny version of SVTR, named SVTR-Tiny. However, SVTR-Tiny is 10 times slower than the recognizer of PP-OCRv2 based on CRNN on CPU with MKLDNN enabled due to the limited model structure supported by the MKLDNN acceleration library, which is not practical enough. As shown in Figure \ref{SVTR_Tiny}, the main structure in SVTR-Tiny is Mix Block, which is proved to be the most time-consuming module through analysis, so we optimize the structure in three steps to speed up and ensure the effectiveness of the model, as shown in Figure \ref{svtr_modify_process}.

Firstly, considering the high efficiency of PP-LCNet, we replace the first half network of SVTR-Tiny with the first three stages of PP-LCNet, and retain only 4 Global Mix Blocks. Secondly, we further reduce the number of Global Mix Blocks from 4 to 2. Thirdly, as we found the prediction speed of Global Mix Block is related to the shape of input features, we move Global Mix Block behind the pooling layer. Finally, we get a novel lightweight text recognition network SVTR-LCNet, which is shown in Figure \ref{svtr_modify_process} (c).

\subsubsection{GTC: Guided Training of CTC by Attention}
Connectionist Temporal Classification (CTC) and attention mechanism are two main approaches used in recent scene text recognition works. Compared with attention-based methods, CTC decoder can achieve a much faster prediction speed, but lower accuracy. To obtain an efficient and effective model, we use an attention module to guide the training of CTC to fuse multiple features, referring to the GTC \cite{Hu2020GTCGT} method, which is effective for the improvement of accuracy. As the attention module is completely removed during prediction, no more time cost is added in the inference process.

\subsubsection{TextConAug: Data Augmentation for Mining Text Context Information}
TextConAug is a data augmentation strategy for mining textual context information. The main idea comes from the paper ConCLR \cite{zhang2022context}, in which the author proposed data augmentation strategy ConAug to concat 2 different images in a batch to form new images and perform self-supervised comparative learning. We apply this method to supervised learning tasks, and design TextConAug which can enrich the context information of training data and improve the diversity of training data.

\subsubsection{TextRotNet: Self-Supervised Pre-trained Model}
TextRotNet is a pre-trained model trained with a large amount of unlabeled text line data in a self-supervised manner, referred to previous work STR-Fewer-Labels \cite{baek2021STRfewerlabels}. We use this model to initialize the weights of SVTR-LCNet, helping the text recognition model to converge better. 

\subsubsection{U-DML: Unified-Deep Mutual Learning}
U-DML is a strategy proposed in PP-OCRv2 which is very effective to improve the accuracy without increasing model size. In PP-OCRv3, for two different structures SVTR-LCNet and attention module, the feature map of PP-LCNet, the output of the SVTR module and the output of the Attention module between them are simultaneously supervised and trained.

\subsubsection{UIM: Unlabeled Images Mining}
UIM is a simple unlabeled data mining strategy. The main idea is to use a high-precision text recognition model to predict unlabeled images to obtain pseudo-labels, and select samples with high prediction confidence as training data for training lightweight models.

\section{Experiments}
\subsection{Experimental Setup}
\subsubsection{DataSets}
We perform experiments on the datasets as shown in Table \ref{dataset_ch}, which is expanded on the basis of what we used in our previous work PP-OCR \cite{du2020pp} and PP-OCRv2 \cite{du2021pp}.

For text detection, there are 127k training images and 200 validation images. The training images consist of 68K real scene images and 59K synthetic images. The real scene images are collected from  Baidu image search and public datasets, including LSVT \cite{sun2019chinese}, RCTW-17 \cite{shi2017icdar2017}, MTWI 2018 \cite{mtwi}, CASIA-10K \cite{he2018multi}, SROIE \cite{huang2019icdar2019}, MLT 2019 \cite{nayef2019icdar2019}, BDI \cite{karatzas2011icdar}, MSRA-TD500 \cite{yao2012detecting} and CCPD 2019 \cite{xu2018towards}. The synthetic images mainly focus on the scenarios for long texts, multi-direction texts and texts in table. The validation images are all from real scenes.

For text recognition, there are 18.5M training images and 18.7K validation images. Among the training images, 7M images are real scene images, which come from some public datasets and Baidu image search. The public datasets include LSVT, RCTW-17, MTWI 2018, CCPD 2019, openimages \url{https://github.com/openimages/dataset} and InvoiceDatasets \url{https://github.com/FuxiJia/InvoiceDatasets}. Besides, we scraped 750k financial report images from the web. We get 810k images from LSVT unlabeled data by using UIM strategy. We also obtain about 3M croped images from Pubtabnet \url{https://github.com/ibm-aur-nlp/PubTabNet}.
The remaining 11.5M synthetic images mainly focus on scenarios for different backgrounds, rotation, perspective transformation, noising, vertical text, etc. The corpus of synthetic images comes from the real scene images. All the validation images also come from the real scenes.

\begin{figure}[t]
\centering
\subfigure{
\centering
\includegraphics[width=\columnwidth]{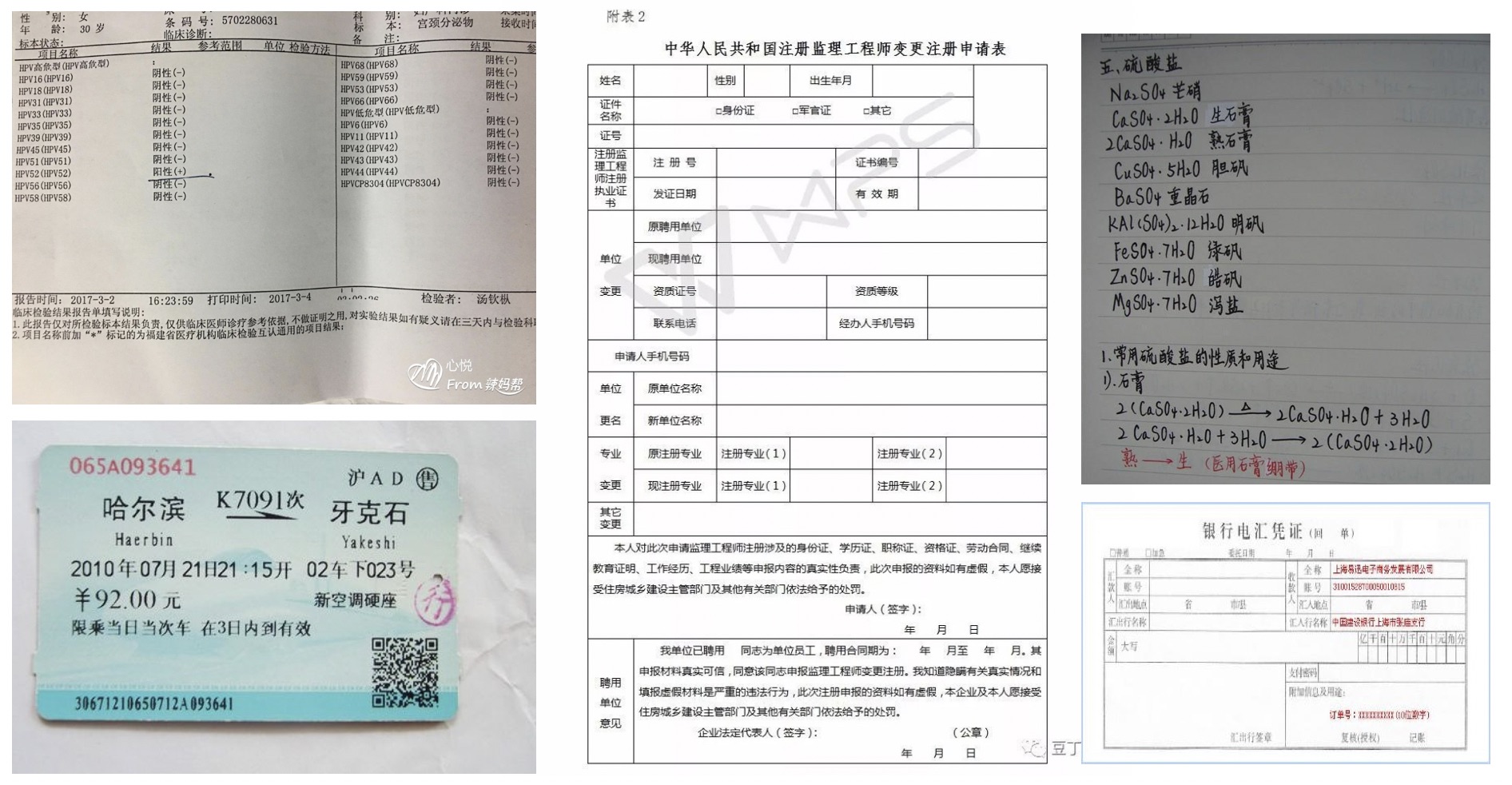}
}
\caption{Some images in the end-to-end test set.}
\label{doc}
\end{figure}

In addition, we collected 800 images for different real application scenarios to evaluate the overall OCR system, including contract samples, license plates, nameplates, train tickets, test sheets, forms, certificates, street view images, business cards, digital meter, etc. Figure \ref{doc} shows some images of the test set.

The data synthesis tool used in text detection and text recognition is modified from text render \cite{textrender}.
    
\subsubsection{Implementation Details}
We adopt most of the strategies used in PP-OCRv2, as you can found in Figure \ref{framework}. We use Adam optimizer to train all the models, setting the initial learning rate to 0.001. The difference is that we adopt cosine learning rate decay as the learning rate schedule for the training of detection model, but piece-wise decay for recognition model training. Besides, we use weight decay 3e-5 for recognition model but 1e-5 for CTC Head. For detection model training, we use 5e-5 for weight decay.
Warm-up training for a few epochs at the beginning is utilized for both detection and recognition models training.

For text detection, the model is trained for 500 epochs in total with warm-up training for 2 epochs. The batch size is set to 8 per card. For text recognition, the model warm up for 5 epochs and is then trained for 700 epochs with the initial learning rate 0.001, and then trained for 100 epochs with learning rate decayed to 0.0001. The batch size is 128 per card.

In the inference period, Hmean is used to evaluate the performance of the text detector and the end-to-end OCR system. Sentence Accuracy is used to evaluate the performance of the text recognizer. GPU inference time is tested on a single T4 GPU. CPU inference time is tested on a Intel(R) Xeon(R) Gold 6148.

\begin{table*}[htbp]
\begin{center}
\resizebox{0.75\textwidth}{!}{
\begin{tabular}{c|c|c|c|c}
\hline
& \multicolumn{3}{c|}{Number of training data} & \makecell[c]{Number of validation data} \\
\cline{2-5}
Task & Total & Real & Synthesis & Real \\
\hline
Text Detection & 127K & 68K & 59K & 500  \\
Text Recognition & 18.5M & 7M & 11.5M & 18.7K  \\
\hline
\end{tabular}}
\end{center}
\caption{Statistics of datasets for text detection and recognition.}
\label{dataset_ch}
\end{table*}

\subsection{Text Detection}
PP-OCRv3 text detector adopts CML distillation strategy which involves a teacher model and two student models as shown in Figure \ref{CML} . We firstly optimize the network of the teacher model and the student model respectively, and then use the optimized teacher model to guide the training of the student model. For the teacher model, LK-PAN is integrated and DML is adopted to further improve the effectiveness. For the student model, RSE-FPN is integrated.

The ablation study is shown in Table \ref{ablation_det}. The table can be divided into two parts according to the double horizontal lines. The upper part is the experimental results of the teacher model, and the lower part is the experimental results of the student model. The teacher model is for better effect and does not consider efficiency while the student model needs to consider both.

\begin{table*}[htbp]
\begin{center}
\resizebox{0.85\textwidth}{!}{
\begin{tabular}{c|l|c|c|c}
\hline
ID & Strategy & \makecell{Model \\ Size(M)} & Hmean(\%) & Speed(ms) \\
\hline
baseline teacher & DB-R50 & 99 & 83.5 & 260 \\
teacher1 & DB-R50-LK-PAN & 124 & 85.0 & 396 \\
teacher2 & DB-R50-LK-PAN-DML & 124 & \textbf{86.0} & 396 \\
\midrule
\specialrule{0em}{1.5pt}{1.5pt}
\midrule
student1* & DB-MV3  & 3 & 81.3 & 117 \\
student2* & DB-MV3-RSE-FPN & 3.6 & 84.5 & 124 \\
baseline student & PP-OCRv2 det & 3 & 83.2 & 117 \\
student1 & DB-MV3-CML(teacher2) & 3 & 84.3 & 117 \\
student2 & DB-MV3-RSE-FPN-CML(teacher2) & 3.6 & \textbf{85.4} & 124 \\
\hline
\end{tabular}}
\end{center}
\caption{Ablation study of enhancement strategies for text detection. DB-R50 means DB text detection model with ResNet50 backbone. DB-R50-LK-PAN replaces DB-R50's FPN module with LK-PAN. DB-R50-LK-PAN-DML is DB-R50-LK-PAN trained with DML. DB-MV3 means DB model with MobileNetV3 backbone. DB-MV3-RSE-FPN replaces DB-MV3's FPN module with RSE-FPN. DB-MV3-CML is distilled from DB-R50-LK-PAN-DML. DB-MV3-RSE-FPN-CML is distilled from DB-R50-LK-PAN-DML with RSE-FPN integrated.}
\label{ablation_det}
\end{table*}

The baseline of the teacher model is a DB model with a backbone of ResNet50, named DB-R50. It can be found that Hmean can be improved from 83.5\% to 85.0\% by using LK-PAN, with the inference time cost increasing from 260ms to 396ms. Using DML distillation, Hmean of teacher model can be further improved to 86.0\%. 

For student model, comparing the two experiments marked with *, which means experiments without CML distilling method, we can find that the Hmean can be improved from 81.3\% to 84.5\% with RSE-FPN, with the inference time increasing by only 6\%. Student1* is equivalent to PP-OCR mobile detector, while PP-OCRv2 detector is a distilled version of PP-OCR mobile detector by CML. 

\begin{figure*}[htbp]
\centering
\subfigure{
\centering
\includegraphics[width=\textwidth]{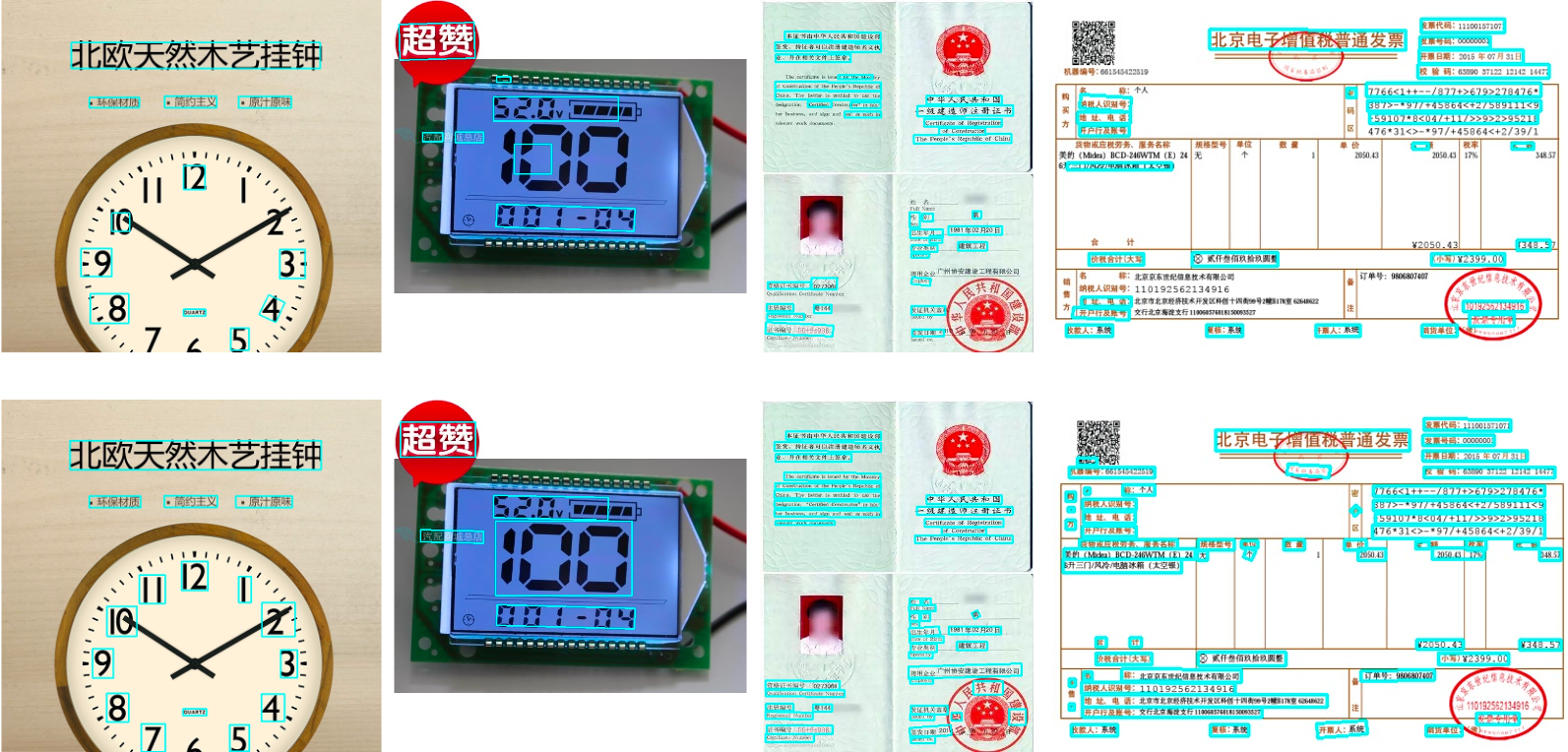}
}
\caption{Comparison between PP-OCRv2 and PP-OCRv3 text detection models. The upper and lower rows are from PP-OCRv2 and PP-OCRv3 respectively.}
\label{vis_det}
\end{figure*}

Furthermore, we verify the effectiveness of the combination of the optimized teacher and student models in CML. DB-MV3-CML is trained with CML, guided by the teacher model DB-R50-LK-PAN-DML. The Hmean is improved from 83.2\% to 84.3\%. If we introduce RSE-FPN in the student model of DB-MV3-CML, the Hmean can be improved from 84.3\% to 85.4\%. Finally, we adopt DB-MV3-RSE-FPN-CML as the text detection model of PP-OCRv3. The visual comparison of PP-OCRv2 and PP-OCRv3 text detection model is shown in Figure \ref{vis_det}.

\subsection{Text Recognition}
Table \ref{Ablation_SVTR} shows the ablation study of SVTR-LCNet. We choose PP-OCRv2 baseline model as our baseline, which use PP-LCNet, BiLSTM with hidden size 48 and CTC decoder but U-DML. Comparing SVTR-Tiny with PP-OCRv2 baseline, the accuracy can be improved by 10.8\%, while the prediction speed nearly 11 times slower. After replacing the first half network of SVTR-Tiny with the first three stages of PP-LCNet, retaining only 4 Global Mix Blocks, the accuracy is reduced to 76\%, but the speed can be increased by 69\%. Then we further reduce the number of Global Mix Blocks from 4 to 2, which further reduce the accuracy to 72.9\%, but increase the speed by another 69\%. After moving the Global Mix Block behind the pooling layer, the accuracy drops to 71.9\%, and the prediction speed surpasses the CNN-based PP-OCRv2 baseline by 22\%. In addition, the height of the input image is further increased from 32 to 48, which makes the prediction speed slightly slower, but the accuracy greatly improved. Using SVTR-LCNet, the recognition accuracy reaches 73.98\%, which is close to the accuracy of PP-OCRv2 recognizer trained with U-DML distillation.

\begin{table}[tb]
\begin{center}
\begin{tabular}{c|c|c|c}
\hline
Strategy & \makecell{Model \\ Size(M)} & Acc(\%) & Speed(ms) \\
\hline
PP-OCRv2 baseline & 8 & 69.3 & 8.5 \\
SVTR-Tiny & 21 & 80.1 & 97.0 \\
SVTR-LCNet(G4) & 9.2 & 76.0 & 30.0 \\
SVTR-LCNet(G2) & 13 & 73.0 & 9.4 \\
SVTR-LCNet(h32) & 12 & 71.9 & 6.6 \\
SVTR-LCNet(h48) & 12 & 74.0 & 7.6 \\

\hline
\end{tabular}
\end{center}
\caption{Ablation study of SVTR-LCNet. G4 means 4 Global Mix Blocks and G2 means 2 Global Mix Blocks. h32 means the height of input image is 32pixel, and h48 means the height of input image is 48pixel. The prediction speed is tested on CPU.}
\label{Ablation_SVTR}
\end{table}

\begin{figure*}[htbp]
\centering
\subfigure{
\centering
\includegraphics[width=0.8\textwidth]{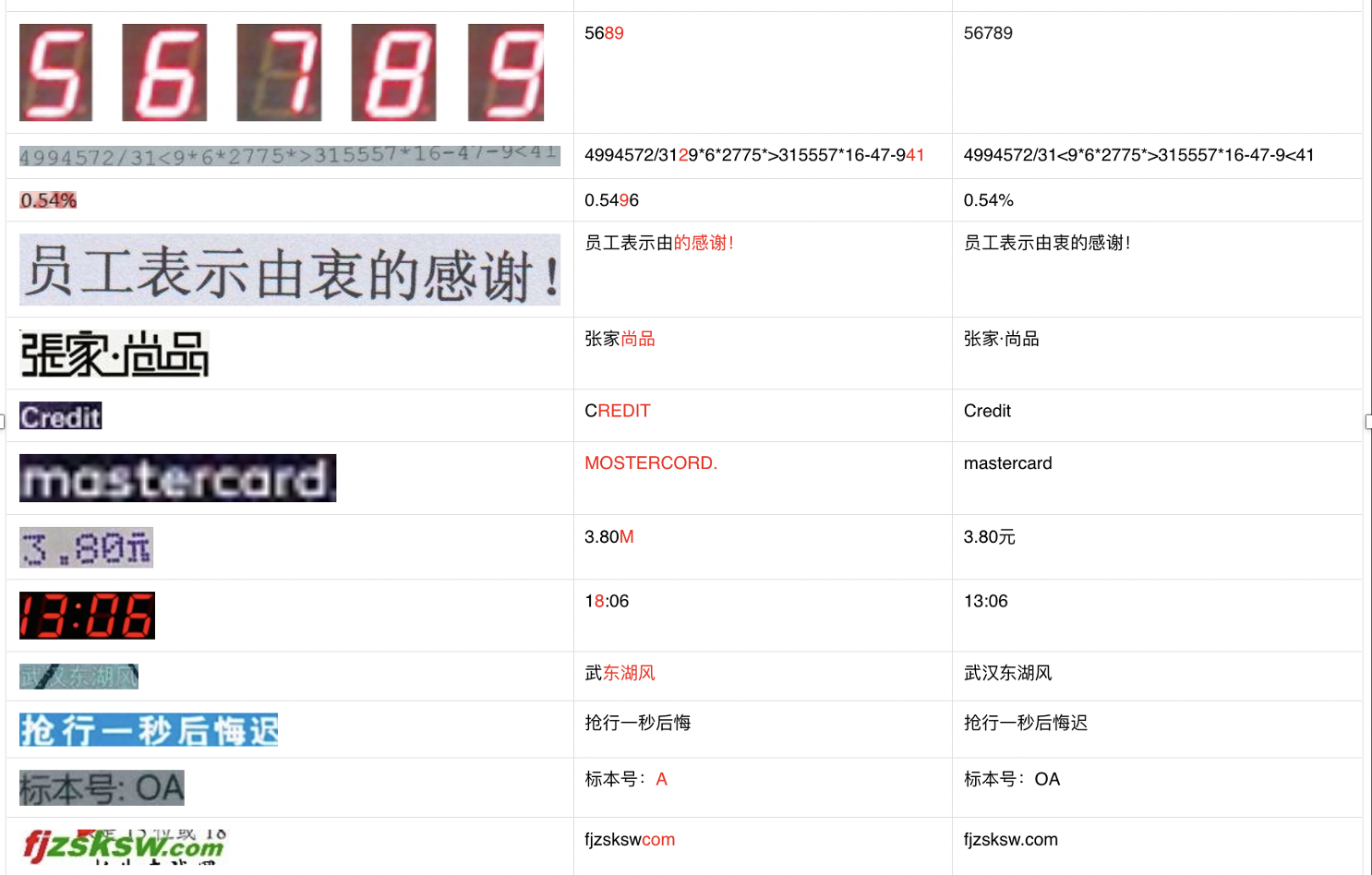}
}
\caption{Comparison between PP-OCRv2 and PP-OCRv3 text recognition models. The left column shows test images, the middle column shows test results of PP-OCRv2, the right column shows test results of PP-OCRv3.}
\label{rec_examples}
\end{figure*}

Table \ref{Ablation_all} shows the ablation study of optimization strategies for PP-OCRv3 recognizer. Comparing SVTR-LCNet with PP-OCRv2, the accuracy of SVTR-LCNet is close to the accuracy of PP-OCRv2 recognizer trained with U-DML distillation, and the speed is increased by 11\%. GTC can improve the accuracy by 1.82\%, and no more time-consuming is added in the inference process as the attention module is completely removed during prediction. By using TextConAug, the accuracy is further improved by 0.5\%. TextRotNet can improve the accuracy by another 0.6\%. Furthermore, the accuracy can be improved by 1.5\% by U-DML, which is a significant improvement. By using UIM to mine unlabeled data, the accuracy can be improved by another 1.0\%. Figure \ref{rec_examples} show some examples tested by PP-OCRv3 and PP-OCRv2 recognizer.

\begin{table}[tb]
\begin{center}
\begin{tabular}{c|c|c|c}
\hline
Strategy & \makecell{Model \\ size(M)} & Acc(\%) & Speed(ms) \\
\hline
PP-OCRv2 & 8 & 74.8 & 8.5 \\
SVTR-Tiny & 21 & 80.1 & 97.0 \\
SVTR-LCNet(h32) & 12 & 71.9 & 6.6 \\
SVTR-LCNet(h48) & 12 & 74.0 & 7.6 \\
+ GTC & 12 & 75.8 & 7.6 \\
+ TextConAug & 12 & 76.3 & 7.6 \\
+ TextRotNet & 12 & 76.9 & 7.6 \\
+ UDML & 12 & 78.4 & 7.6 \\
+ UIM & 12 & \textbf{79.4} & 7.6 \\

\hline
\end{tabular}
\end{center}
\caption{Ablation study of PP-OCRv3 recognition. + means a new strategy is used based on previous strategies. The prediction speed is tested on CPU.}
\label{Ablation_all}
\end{table}

\subsection{System Performance}
In Table \ref{end2end}, we compare the performance between proposed
PP-OCRv3 with previous ultra lightweight PP-OCR systems. As we can see, the Hmean of PP-OCRv3 is 5.3\% higher than that of PP-OCRv2 with the same inference cost on CPU. The inference speed of PP-OCRv3 is 22\% faster than PP-OCRv2 on T4 GPU. The visualized results PP-OCRv3 is shown in Figure \ref{vis_sys}.

\begin{table}[tb]
\begin{center}
\begin{tabular}{c|c|c|c|c}
\hline
& & & \multicolumn{2}{c|}{Speed(ms)} \\
\cline{4-5}
Model & Hmean(\%) & \makecell{Model \\ size(M)} & CPU & T4 GPU \\
\hline
\makecell{PP-OCR \\ mobile} & 50.3 & 8.1 & 356 & 116 \\
\makecell{PP-OCR \\ server} & 57.0 & 155.1 & 1056 & 200 \\
\makecell{PP-OCR \\ v2} & 57.6 & 11.6 & 330 & 111 \\
\makecell{PP-OCR \\ v3} & \textbf{62.9} & 15.6 & 331 & \textbf{87} \\

\hline
\end{tabular}
\end{center}
\caption{Comparison between PP-OCRv3 system and previous PP-OCR systems.}
\label{end2end}
\end{table}

\section{Conclusions}
In this paper, we propose a more robust OCR system PP-OCRv3 which involves 9 improvements, 3 of which are for the detector and 6 for the recognizer. Experiments demonstrate that the Hmean of PP-OCRv3 outperforms PP-OCRv2 by 5\% with the same prediction cost. The corresponding ablation experiments are also provided.

\bibstyle{aaai21}
\bibliography{eg}

\end{document}